\documentclass[11pt]{article}

\usepackage[preprint]{acl}

\usepackage{times}
\usepackage{latexsym}

\usepackage[T1]{fontenc}

\usepackage[utf8]{inputenc}

\usepackage{microtype}

\usepackage{inconsolata}

\usepackage{graphicx}
\usepackage{amsmath} 
\usepackage{booktabs}
\usepackage{multirow}
\usepackage{booktabs} 
\usepackage{multirow} 
\usepackage{tabularx} 
\usepackage{subcaption}
\usepackage{float}
\usepackage{amsmath, amssymb}

\usepackage[table]{xcolor} 
\definecolor{myblueBase}{HTML}{BBE0EF}
\colorlet{myblue}{myblueBase!30}
\usepackage{tcolorbox}
\usepackage{xcolor}
\usepackage{enumitem}

\title{MCite-RL: Towards Reliable Multimodal RAG via Citation-enhanced Agentic Reinforcement Learning}

\author{
 \textbf{Suifeng Zhao\textsuperscript{1}},
 \textbf{Zida Liu\textsuperscript{1}},
 \textbf{Xinyu Lei\textsuperscript{2}},
 \textbf{Lei Sun\textsuperscript{3}},
 \textbf{Jun Gao\textsuperscript{1*},
 \textbf{Sujian Li\textsuperscript{2}}}\thanks{Corresponding Authors}\\
 \textsuperscript{1}Key Laboratory of High Confidence Software Technologies, CS, Peking University, China\\
 \textsuperscript{2}Key Laboratory of Computational Linguistics(MOE), CS, Peking University, China\\
 \textsuperscript{3}Panasonic Connect Co., Ltd.
Tokyo, Japan
 \\
 \texttt{\{sfzhao25,zdliu25,xinyulei25\}@stu.pku.edu.cn},\\
 \texttt{sun.lei@jp.panasonic.com},\\
\texttt{\{gaojun,lisujian\}@pku.edu.cn}
}

\begin{document}
\maketitle
\begin{abstract}
Multimodal Retrieval-Augmented Generation (RAG) with visual citation is crucial for ensuring the traceability and verifiability of MLLMs. However, current RAG and SFT-based methods struggle to achieve robust cross-modal reasoning, causing imprecise visual citations or decoupling between the citation and the generated answers.
To address these limitations, we propose MCite-RL, a citation-enhanced agentic reinforcement learning framework designed for reliable multimodal RAG. MCite-RL introduces an Agentic Refinement module for visual citation that employs iterative retrieval, reasoning, and recursive cropping to progressively narrow the search space, transforming citation into a dynamic, evidence-driven reasoning process rather than a static step. Furthermore, we incorporate a Citation-enhanced Reward mechanism that integrates both process-level and outcome-level feedback within a reinforcement learning paradigm to jointly optimize answer accuracy and source traceability. Extensive experiments on benchmarks such as Wiki-VISA, FinRAGBench-V, and MMLongBench-Doc demonstrate that MCite-RL effectively achieves the joint optimization of citation precision and answer quality.
\end{abstract}

\section{Introduction}

\begin{figure}[t]
  \centering
  \includegraphics[width=\columnwidth]{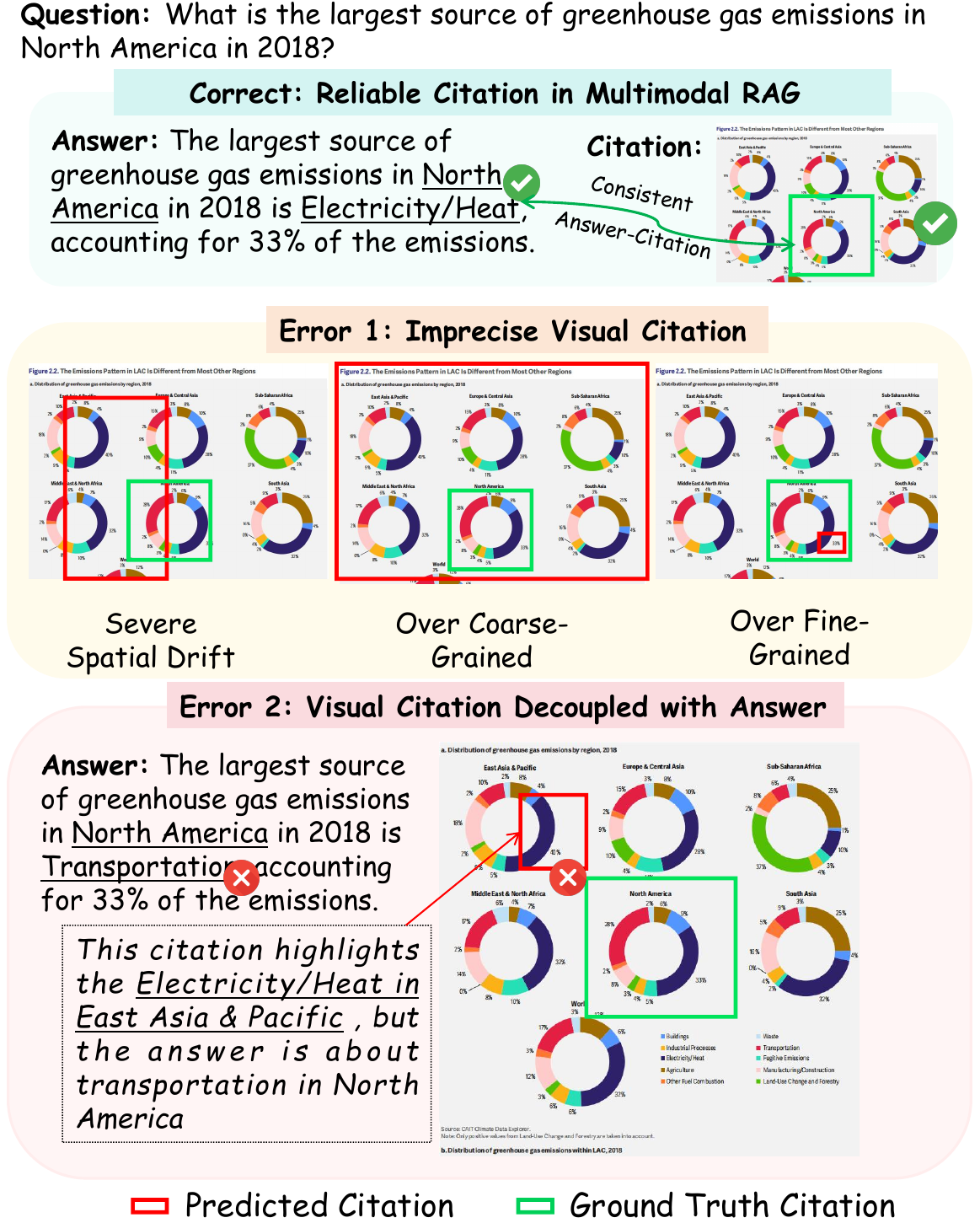}
 \caption{Typical success and failure patterns of visual citation in multimodal RAG. A reliable case (top) exhibits precise visual citation that directly supports the generated answer. In contrast, existing models frequently suffer from imprecise citations (middle) or scenarios where citations become decoupled from the generated answer (bottom).}
  \label{fig:intro}
\end{figure}

Recent advances in Multimodal Large Language Models (MLLMs) have enabled Retrieval-Augmented Generation (RAG) systems \citep{gao2024retrievalaugmentedgenerationlargelanguage, jin-etal-2023-instructor, wei2025instructraginstructingretrievalaugmentedgeneration, Yoran2023MakingRL} to move beyond text-only settings and incorporate visual information into the generation process, allowing models to produce richer and more informative responses \citep{wang2025vidoragvisualdocumentretrievalaugmented, yu2025visragvisionbasedretrievalaugmentedgeneration, suri-etal-2025-visdom}. Similar to text-based RAG systems \citep{DBLP:conf/emnlp/GaoYYC23,DBLP:journals/corr/abs-2601-04525,DBLP:journals/corr/abs-2601-06021}, 
accurate and well-grounded citations are critical for multimodal RAG \citep{DBLP:journals/corr/abs-2510-09733}, as they improve the verifiability and traceability of generated outputs by explicitly linking responses to the underlying visual evidence.

Multimodal RAG with Visual Citation has recently attracted initial research interest as a framework  aimed at enhancing both accuracy and verifiability. Initial efforts have primarily focused on defining the task and establishing preliminary benchmarks. For instance, VISA \cite{DBLP:conf/acl/MaZKZCL25} pioneers the concept of visual source attribution by linking answers to fine-grained visual evidence through SFT-based methods, and FinRAGBench-V \cite{DBLP:journals/corr/abs-2505-17471}, which extends this task by introducing automated evaluation for visual citations in complex documents. For these benchmarks, the adopted baseline methods are still relatively simple, typically relying on vanilla RAG or SFT-based approaches to implement basic citation capabilities.


Despite these emerging efforts, we argue that multimodal RAG with visual citation entails more fundamental challenges that remain insufficiently explored. \textbf{First, there remains a substantial gap in citation precision}. This difficulty stems from the need to localize relevant evidence at an appropriate level of granularity, where both overly coarse and overly fine-grained citation can lead to incorrect visual support~\citep{DBLP:journals/corr/abs-2601-08620,DBLP:conf/acl/MaZKZCL25,DBLP:journals/corr/abs-2511-15090}. As illustrated in Figure~\ref{fig:intro}, MLLMs frequently fail to localize the correct visual evidence, generating bounding boxes that are spatially misaligned, too coarse, or too detailed. Such imprecise citations fail to serve as verifiable evidence. 
\textbf{Second, 
visual citations are often misaligned with the generated answers they are meant to support}. 
As shown in the bottom example of Figure~\ref{fig:intro}, the model may match a numerical value yet cite a visual region from an entirely different context.  This mismatch reveals that the answer and the citation lack a shared evidence basis, fundamentally undermining the trustworthiness of citation-enabled multimodal RAG systems.

Based on the above analysis, we conclude that a reliable multimodal RAG system requires far more beyond accurate answer, it also requires precise evidence localization and consistent answer-citation alignment. However, achieving such fine-grained citation in complex multimodal documents is inherently difficult within a static, single-step generation paradigm~\citep{DBLP:journals/corr/abs-2601-06021,DBLP:journals/corr/abs-2510-09733}. Existing approaches, whether based on prompting or static supervised fine-tuning, often fail to provide the appropriate granular feedback necessary to align these objectives. We require both a holistic feedback mechanism to ensure citation reliability and an iterative agentic workflow to progressively refine visual evidence. Motivated by this insight, we propose \textbf{MCite-RL}, a novel citation-enhanced agentic reinforcement learning framework. Unlike previous methods, we explicitly treat citation precision as a primary optimization objective by incorporating it as a dedicated component in the reward function, rather than considering it treating it merely as a by-product of answer generation. Inspired by prior work on iterative retrieval and reasoning (e.g., Search-R1 \cite{DBLP:journals/corr/abs-2503-09516}; VRAG-RL \cite{DBLP:journals/corr/abs-2505-22019}), we also design an agentic refinement workflow for visual citation, which progressively narrows the visual search space through multi-step evidence refinement, allowing answers and their corresponding citations to emerge from a shared, evidence-driven reasoning process.

We evaluate the effectiveness of our method on three benchmarks, including Wiki-VISA \cite{DBLP:conf/acl/MaZKZCL25}, FinRAGBench-V \cite{DBLP:journals/corr/abs-2505-17471}, and MMLongBench-Doc \cite{ma2024mmlongbenchdocbenchmarkinglongcontextdocument}. Experimental results show that MCite-RL consistently improves citation precision across all benchmarks, achieving an average gain of 26.16\% on the Intersection-over-Union (IoU) metric compared to the vanilla RAG baseline. Moreover, the integration of citation-based rewards also leads to an average increase of 5.89\% in answer accuracy, demonstrating that explicit citation supervision directly contributes to answer reliability.

Our contributions are summarized as follows:
\begin{itemize}
    \item We propose MCite-RL, a framework that improves the reliability of multimodal RAG by integrating citation-enhanced agentic reinforcement learning. 
    
    \item We introduce an agentic workflow for multimodal RAG with visual citation, where iterative evidence refinement enables answers and citations to emerge from a shared reasoning process.
    
    \item We develop a citation-enhanced reward function that explicitly optimizes both reasoning accuracy and citation precision. By incorporating citation feedback at multiple levels, our method leverages citation precision as a central driver for model training.
    
    \item Extensive experiments demonstrate the effectiveness of our approach, showing improvements not only in citation precision but also in answer accuracy.
\end{itemize}
\section{Related Work}

\paragraph{Multimodal RAG with Visual Citation}
Early Multimodal RAG systems primarily focused on coarse-grained retrieval over documents or images, aiming to augment generation with multimodal context \citep{xia2025mmedragversatilemultimodalrag, faysse2024colpaliefficientdocumentretrieval, ma2024unifyingmultimodalretrievaldocument}. More recent studies have shifted toward visual citation, which requires grounding generated content in fine-grained image regions to enable traceability and verification. VISA \cite{DBLP:conf/acl/MaZKZCL25} formalized this setting through visual source attribution by aligning generated text spans with detected visual regions, while FinRAGBench-V \cite{DBLP:journals/corr/abs-2505-17471} systematically evaluated visual citation performance in visually dense financial charts and tables. Despite these advances, most existing approaches rely on static and passive grounding pipelines, where visual regions are pre-selected prior to generation and loosely aligned afterward, limiting citation precision and robustness in complex multimodal reasoning scenarios.

\paragraph{Agentic Reasoning and Reinforcement Learning}
The rise of agentic Large Language Models has transformed RAG from one-shot retrieval into iterative reasoning–action loops \citep{Xie2025InterleavedRF, Wang2025GenR1SearcherCR,Tan2025RAGR1IT}, exemplified by the ReAct paradigm \cite{yao2023reactsynergizingreasoningacting} and subsequent efforts on agentic reasoning \citep{wu-etal-2025-agentic, tongyidr, kimiresearcher, li2025search,Zhang2025LeTSLT,Jiang2025QAgentAM,Shi2025SearchAR,Jiang2025s3YD,Wu2025StructureR1DL,Wang2025StepSearchIL}. In text-only setting, Search-R1 \cite{DBLP:journals/corr/abs-2503-09516} employs chain-of-thought reasoning to refine retrieval queries, while in multimodal scenarios, VRAG-RL \cite{DBLP:journals/corr/abs-2505-22019} incorporates reinforcement learning to iteratively crop and attend to visual regions. However, these methods predominantly optimize answer correctness or task-level rewards, treating visual grounding as an implicit intermediate signal rather than an explicit optimization target. Consequently, models may internally attend to relevant regions without producing precise, verifiable visual citations.

\begin{figure*}[t]
  \centering
  \includegraphics[clip=true,width=1\textwidth]{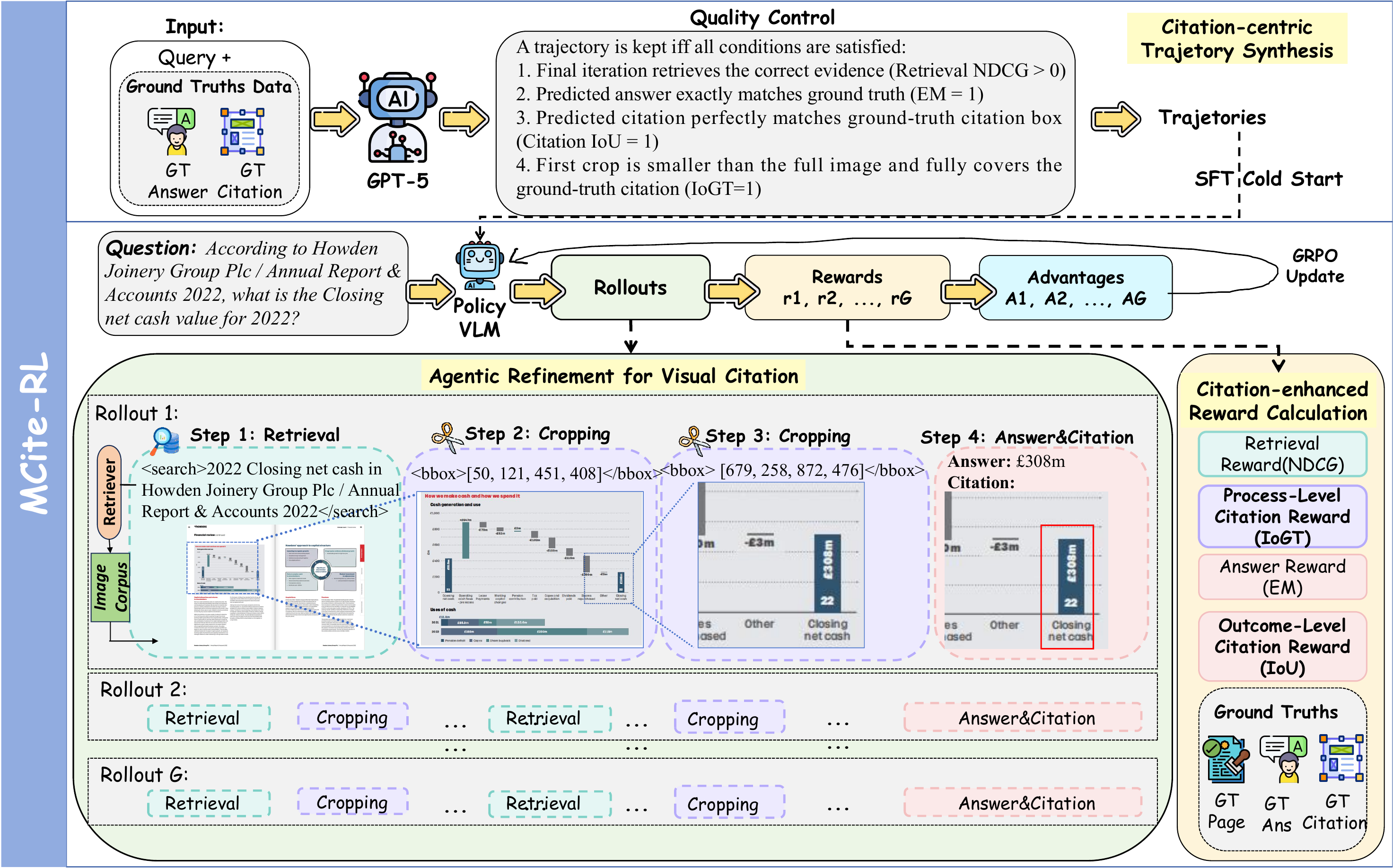}
 \caption{\textbf{Architecture of MCite-RL.} It features a two-stage pipeline: (1) Trajectory Synthesis with rigorous quality control for supervised fine-tuning (SFT), and (2) RL-based Agentic Refinement for visual citation, which leverages multi-step rollouts (retrieval and cropping) and citation-enhanced rewards to jointly optimize answer accuracy and citation precision.}
  \label{fig:MCite-RL}
\end{figure*}
\section{MCite-RL}

Figure~\ref{fig:MCite-RL} illustrates the overall architecture of MCite-RL, which consists of two core components: an agentic refinement module for visual citation (Section \ref{sec:agentic}) and a citation-enhanced reward calculation module (Section \ref{sec:citation-enhanced rl}). Based on this architecture, the model is trained via a two-stage process (Section~\ref{sec:training_method}) comprising SFT Cold-Start and Citation-enhanced RL.

\subsection{Task Formulation of Multimodal RAG with Citation}

We formally define the task of Multimodal Retrieval-Augmented Generation with Citation (MMRAG with Citation) as a constrained generation problem. Given a user query $q$ and an initial image $i$, the objective is to generate a response tuple $(y, c_{final})$, where $y$ represents the textual answer and $c_{final} = \{(x_1, y_1), (x_2, y_2)\}$ denotes the visual citation in the form of bounding box coordinates within $i$ that provide precise visual evidence for the answer $y$. The overall process can be formulated as a function: 
\begin{equation}
    \mathcal{F}: (q, i) \to (y, c_{final}).
\end{equation}

To effectively navigate the vast search space and ensure grounding precision, the generation process is decomposed into the synergistic interplay between iterative retrieval and progressive visual cropping. Unlike standard VQA, this task requires the joint optimization of two critical objectives: \textbf{answer accuracy}, which measures the fidelity of the answer $y$ to the query $q$, and \textbf{citation precision}, which ensures the spatial alignment between the citation $c_{final}$ and the corresponding segments in $y$. Since optimizing $\mathcal{F}$ involves mastering the interplay between reasoning and tool manipulation, we adopt an agentic refinement workflow to manage this iterative process.

\subsection{Agentic Workflow for Visual Citation}\label{sec:agentic}

As illustrated in lower-left part of Figure~\ref{fig:MCite-RL}, we employ an agentic reasoning pipeline to MMRAG outputs with precise visual citations. 
While iterative retrieval and cropping have been explored to enhance general QA performance \citep{DBLP:journals/corr/abs-2503-09516, DBLP:journals/corr/abs-2505-22019,Mei2025AISearchPlannerMA},
we adapt this workflow to enforce both answer accuracy and citation precision. 

Within our framework, this iterative process progressively collects the visual and textual clues needed to answer the question. By alternating between image retrieval and visual grounding, the agent successively crops the visual field by pruning irrelevant regions, thereby narrowing its focus onto the precise coordinates required to anchor the final answer.

\noindent\textbf{Iterative Visual Grounding.}
We formulate inference as a sequential decision-making trajectory:
\begin{equation}
    \tau = \{(s_0, a_0), (s_1, a_1), \dots, (s_T, a_T)\},
\end{equation}
where $s_t$ denotes the agent's state at step $t$ and $a_t$ denotes the corresponding action. The action space includes image retrieval, visual cropping, and generating the final answer with visual citations.

Each state $s_t$ is represented as:
\begin{equation}
    s_t = (q, h_t, v_{c_t}),
\end{equation}
where $q$ is the input query, $h_t$ summarizes the interaction history up to step $t$ (encompassing past actions and retrieved contexts), and $v_{c_t}$ is the visual observation, which is cropped based on the generated coordinates $c_t$ at step $t$.

As shown in the lower-left part of Figure \ref{fig:MCite-RL}, the agent progressively refines its visual observation through iterative retrieval and cropping. The goal of this refinement is to identify an optimal visual region $\hat{v}_{c_t}$. This region is selected to maximize the likelihood of the evidence given the interaction history and the query:
\begin{equation}
    \hat{v}_{c_t} = \arg\max_{v} P(v \mid \tau_{<t}, q)
\end{equation}
where $\tau_{<t} = \{(s_0, a_0), \dots, (s_{t-1}, a_{t-1})\}$ represents the historical trajectory of states and actions.

\paragraph{Joint Answer and Citation Generation.}
Based on the final converged visual region $v_{c_T}$, the model jointly generates the textual answer $y$ and its corresponding visual citation $c_{final}$ in a single step:

\begin{equation}
(\hat{y}, \hat{c}_{final}) = \arg\max_{y, c_{final}} P(y, c_{final} \mid v_{c_T}, q)
\end{equation}
Both the answer and the citation are derived from the shared visual context, with the citation spatially localized within $v_{c_T}$. This joint generation mechanism ensures that the answer is intrinsically grounded in visual evidence, establishing strong alignment between the answer and its supporting region while mitigating the risk of misaligned or hallucinated citations.

\subsection{Citation-enhanced Reward}
\label{sec:citation-enhanced rl}
In MCite-RL, we design a citation-enhanced reward function tailored to the agentic interaction trajectory described in Section~\ref{sec:agentic}. This reward jointly captures two aspects: \textbf{process-level} intermediate grounding feedback and \textbf{outcome-level} citation precision, thereby incentivizing the generation of precise visual citation.

\paragraph{Process-Level Citation Reward.} 
To supervise the dynamic reasoning trajectory, we introduce the process-level reward $R_{\text{cit}}^{\text{proc}}$, which evaluates the quality of the converged search space during the agentic loop. While the agent performs multiple “think-and-crop” iterations and gradually narrow the search space, this reward is finally anchored on the terminal crop $c_T$. We employ the Intersection-over-Ground-truth (IoGT) metric to quantify this reward:
\begin{equation}
    R_{\text{cit}}^{\text{proc}} = \text{IoGT}(c_{T}, c^*) = \frac{|c_{T} \cap c^*|}{|c^*|},
\end{equation}
where $c^*$ denotes the ground-truth citation and $|\cdot|$ represents the region area. Unlike the outcome-level reward that emphasizes precise localization, the process-level IoGT specifically incentivizes the preservation of core evidence throughout the visual grounding refinement process. By providing this supervision signal, the reward penalizes any information loss during the iterative pruning of the visual field. This ensures that the agent maintains a high-recall focus on the ground-truth region, thereby establishing a robust and undistorted evidence foundation for the subsequent generation of the textual answer and its corresponding visual citation.

\paragraph{Outcome-Level Citation Reward.} 

The outcome-level reward $R_{\text{cit}}^{\text{out}}$ evaluates the precision of the final predicted citation $c$ against the ground-truth $c^*$ using Intersection-over-Union (IoU):
\begin{equation}
    R_{\text{cit}}^{\text{out}}(c_{final}, c^*) = \text{IoU}(c_{final}, c^*)=\frac{|c_{final} \cap c^*|}{|c_{final} \cup c^*|}.
\end{equation}
Unlike the process-level citation reward, which emphasizes full coverage of the ground truth in intermediate crops, the outcome-level reward focuses on the accuracy of the final prediction. An ideal final citation should be tightly aligned with the ground truth, capturing all necessary information while excluding irrelevant regions. By enforcing this IoU-based criterion, we aim to make the final citation both accurate and verifiable. Together with the process-level reward, this design guides the agent to progressively hone in on the relevant region during intermediate steps, while ultimately producing a highly precise final citation, thereby reinforcing the structural alignment between the answer and its supporting evidence.

\paragraph{Final Reward Function.} 
To achieve holistic optimization, the final reward function $R(\tau)$ integrates retrieval efficiency, answer accuracy, and citation rewards at both outcome and process levels, along with the format penalty. We define the total reward $R(\tau)$ as follows:
\begin{equation}
\small
    R(\tau) = \mathbb{I}_{\text{v}}(\tau) \cdot (\alpha R_{\text{ret}} + \beta R_{\text{ans}} + \gamma_1 R_{\text{cit}}^{\text{out}} + \gamma_2 R_{\text{cit}}^{\text{proc}})
\label{eq:reward}
\end{equation}
where each component is defined below. All reward components are normalized to the range $[0, 1]$. 
\begin{itemize}[leftmargin=*]
    \item $R_{\text{ret}}$ represents the \textbf{retrieval efficiency reward}. Inspired by 
    \citet{DBLP:journals/corr/abs-2505-22019}, 
    we employ nDCG as the metric, to encourage the agent to acquire the relevant image in the earliest possible steps during iterative retrieval.
    \item $R_{\text{ans}}$ represents the \textbf{answer accuracy reward}, 
    calculated based on Exact Match (EM) between the generated response and the ground-truth. 
    \item $R_{\text{cit}}^{\text{out}}$ and $R_{\text{cit}}^{\text{proc}}$ are the \textbf{citation precision rewards} at the outcome and process levels respectively, as described  above. 
     \item $\mathbb{I}_{\text{v}}(\tau) \in \{0, 1\}$ acts as a \textbf{format constraint}. A rollout trajectory is valid if it covers the required agentic sequence: retrieval, cropping, and final response with answer and visual citation. 
\end{itemize}

\subsection{Training Method}
\label{sec:training_method}

\paragraph{SFT Cold-Start.}
To adapt the model to the required agentic reasoning format and establish a high-quality initial policy for subsequent reinforcement learning, we conduct a cold-start phase using supervised fine-tuning (SFT). As existing multi-modal datasets generally lack explicit reasoning trajectories, we synthesize data that formalizes the chain of thought for both textual answers and visual citations. Specifically, given an input query $q$, the ground-truth answer $y^*$, and golden citation coordinates $c^*$, we leverage GPT-5 to generate step-wise agentic trajectories. These trajectories adhere strictly to a predefined workflow: iterative image retrieval, visual cropping, and final answer generation integrated with visual citations. To ensure data fidelity, we apply rigorous filtering criteria (detailed in Table \ref{tab:filtering_criteria} of Appendix~\ref{appendix:SFT_DATA}) to exclude hallucinated or format-violating samples. This process enables the model to internalize the mandatory structural sequence prior to the RL phase.

\paragraph{Citation-enhanced RL.}
Building upon the SFT-tuned model, we transition to reinforcement learning to further optimize the agent's accuracy and citation precision. We employ the Group Relative Policy Optimization (GRPO) algorithm \cite{DBLP:journals/corr/abs-2402-03300}, which optimizes $\pi_\theta$ by maximizing:
\begin{equation}
\small 
\begin{aligned}
    J_{GRPO}(\theta) &= \mathbb{E} \big[ q \sim P(Q), \{\tau_i\}_{i=1}^G \sim \pi_{\theta_{\text{old}}} \big] \\[-0.2em] 
    & \quad \frac{1}{G} \sum_{i=1}^G \big( \mathcal{L}_{\text{clip}}(\theta) - \beta \mathbb{D}_{KL}(\pi_\theta || \pi_{ref}) \big)
\end{aligned}
\end{equation}
where the advantage $\hat{A}_i$ in $\mathcal{L}_{\text{clip}}$ is computed by normalizing our citation-enhanced reward $R(\tau_i)$ (as defined in Eq.~\ref{eq:reward}) within the group:
\begin{equation}
    \hat{A}_i = \frac{R(\tau_i) - \text{mean}(\{R(\tau_j)\}_{j=1}^G)}{\text{std}(\{R(\tau_j)\}_{j=1}^G)}
\end{equation}
By directly optimizing for these rewards, MCite-RL achieves a synergistic improvement of precise visual citation and answer accuracy.
\newcolumntype{C}{>{\centering\arraybackslash}X}

\begin{table*}[t]
\centering
\footnotesize 
\setlength{\tabcolsep}{4pt} 
\begin{tabularx}{\textwidth}{l CCCCCC} 
    \toprule
    \multirow{2}{*}{\textbf{Method}} & \multicolumn{2}{c}{\textbf{Wiki-Visa}} & \multicolumn{2}{c}{\textbf{FinRAGBench-V}} & \multicolumn{2}{c}{\textbf{MMLongBench-Doc}} \\
    \cmidrule(lr){2-3} \cmidrule(lr){4-5} \cmidrule(lr){6-7}
    & Ans. (\%) & Cit. (\%) & Ans. (\%) & Cit. (\%) & Ans. (\%) & Cit. (\%) \\
    \midrule

    \rowcolor{gray!10} \multicolumn{7}{c}{\textbf{\textit{Proprietary Models}}} \\
    GPT-4o  &31.40& 3.40 & 21.23 & 19.91 & 25.51 & 10.80 \\
    Gemini-2.0-flash & 39.60 & 0.08 & 20.44 & 6.31 & 25.56 & 2.83 \\
    Claude-4.5-sonnet & 12.80 & 0.53 & 34.58 & 15.49 & 30.35 & 16.97 \\
    \addlinespace[4pt] 

    \rowcolor{gray!10} \multicolumn{7}{c}{\textbf{\textit{Qwen2.5-VL-3B-Instruct}}} \\
    Vanilla RAG & 41.20 & 0.46 & \textbf{15.21} & 0.33 & \textbf{27.77} & 2.26 \\
    ReAct & 14.80 & 0.00 & 4.59 & 0.72 & 5.42 & 0.27 \\
    Search-R1 & 7.80 & 1.69 & 1.43 & 1.09 & 3.32 & 3.36 \\
    VRAG-RL (w/SFT) & 48.80 & 27.97 & 10.76 & \underline{25.16} & 24.72 & 19.03 \\
    
    \rowcolor{myblue}\textbf{MCite-RL} & \textbf{51.60}& \textbf{33.45} & \underline{13.27} & \textbf{27.04} & \underline{27.64} & \underline{22.22} \\
    \rowcolor{myblue} \hspace{1em} \textit{-- w/o Proc} & \underline{51.20} & \underline{32.21} & 11.98 & 17.66 & 24.24 & \textbf{22.95} \\
    \rowcolor{myblue} \hspace{1em} \textit{-- w/o RL} & 38.20 & 17.24 & 7.17 & 14.23 & 18.73 & 15.56 \\
    \rowcolor{myblue} \hspace{1em} \textit{-- w/o Training} & 10.20 & 0.00 & 5.02 & 7.48 & 4.97 & 0.55 \\
    \addlinespace[4pt]
    
    \rowcolor{gray!10} \multicolumn{7}{c}{\textbf{\textit{Qwen2.5-VL-7B-Instruct}}} \\
    Vanilla RAG & 40.80 & 1.62  & 17.29 & 6.78  & 31.74 & 12.91 \\
    ReAct & 17.80 & 2.30  & 12.98 & 15.29 & 10.21 & 12.07 \\
    Search-R1 & 48.40 & 1.30  & 10.83 & 12.77 & 29.59 & 13.14 \\
    VRAG-RL (w/SFT) & 54.20 & 20.56 & 17.65 & 15.40 & 31.58 & 12.97 \\ 
    
    \rowcolor{myblue} \textbf{MCite-RL} & \textbf{60.00} & \textbf{36.05} & \textbf{21.81} & \textbf{37.22} & \textbf{35.04}& \textbf{30.12} \\
    \rowcolor{myblue} \hspace{1em} \textit{-- w/o Proc} & \underline{57.60} & \underline{34.50} & \underline{21.31} & \underline{30.22} & \underline{33.77} & \underline{26.40} \\
    \rowcolor{myblue} \hspace{1em} \textit{-- w/o RL} & 45.80 & 18.66 & 15.42 & 20.77 & 23.42 & 19.38 \\
    \rowcolor{myblue} \hspace{1em} \textit{-- w/o Training} & 9.60  & 0.04  & 10.94 & 6.33  & 6.47  & 12.17 \\
    \bottomrule
\end{tabularx}
\caption{Main results of MCite-RL compared with baseline methods on Wiki-Visa, FinRAGBench-V, and MMLongBench-Doc.
NOTE: the proprietary models are only tested under the setting of MCite-RL (w/o Training). \textbf{Bold} and \underline{underline} denote the best and second-best performance, respectively. The \colorbox{myblue}{blue-shaded} rows represent our method and its variants. }
\label{tab:qwen-rag-results}
\end{table*}

\section{Experimental Settings}
\subsection{Datasets}
To evaluate performance across diverse domains, we conduct experiments on subsets of three multimodal RAG benchmarks, specifically filtered single-image samples to evaluate performance across diverse domains:
(i) \textbf{Wiki-VISA} \cite{DBLP:conf/acl/MaZKZCL25}: An open-domain dataset requiring long image pages retrieval from Wikipedia; (ii) \textbf{FinRAGBench-V} \cite{DBLP:journals/corr/abs-2505-17471}: A multimodal financial domain benchmark featuring rich charts and tables; (iii) \textbf{MMLongBench-Doc} \cite{ma2024mmlongbenchdocbenchmarkinglongcontextdocument}: A long-context benchmark where we focus on document understanding tasks. For citation evaluation, Wiki-VISA is assessed using the original annotations; FinRAGBench-V is evaluated on a provided subset with citation ground truth; and MMLongBench-Doc is evaluated on 300 samples manually annotated and verified.
Detailed statistics of datasets are in Appendix~\ref{subsec:data_stats}.

\subsection{Baselines and Model Variants}
To 
evaluate its effectiveness, we compare MCite-RL against three categories of methods:

(i) \textbf{Proprietary Models}: We evaluate state-of-the-art closed-source models (GPT-4o \cite{DBLP:journals/corr/abs-2303-08774}, Gemini-2.0-Flash \cite{DBLP:journals/corr/abs-2507-06261}, Claude-4.5-Sonnet \cite{anthropic2024claude35sonnet}) in a zero-shot setting, using the identical prompting strategy as our ``MCite-RL (w/o training)'' setting.

(ii) \textbf{Reproduced Strong Baselines}: We implement standard paradigms (Vanilla RAG, ReAct) and reproduce strong baselines (Search-R1, and VRAG-RL) using the Qwen2.5-VL-Instruct models (3B and 7B) trained on our specific dataset.

(iii) \textbf{Ablation Variants}: We evaluate MCite-RL and its different training variants (w/o Proc, w/o RL, and w/o Training, etc.) to analyze the individual contributions of each component. Specifically, \textit{w/o Proc} removes the process-level citation rewards, utilizing only the final answer correctness for optimization. \textit{w/o RL} refers to the model derived purely from Supervised Fine-Tuning (SFT) without the subsequent RL training stage. Finally, \textit{w/o Training} denotes the direct inference results from the off-the-shelf base model.
\subsection{Evaluation Metrics}
We evaluate performance along two dimensions: answer accuracy and citation precision.
\paragraph{Answer Accuracy.} 
To ensure a fair comparison, we follow the evaluation protocols defined by each original benchmark,  incorporating both rule-based metrics and model-based semantic judgments. Detailed mappings of metrics to specific datasets are provided in Appendix~\ref{appendix:metrics}.
\paragraph{Citation Precision.} 
To quantify the precision of visual citation, we measure the alignment between predicted bounding boxes and ground-truth annotations using the average Intersection over Union (IoU).

\subsection{Training Details}
\label{sec:impl_details}
We implement our MCite-RL framework using Qwen2.5-VL-3B-Instruct and Qwen2.5-VL-7B-Instruct \cite{qwen2.5-VL} as backbone models, and employ ColQwen \cite{faysse2024colpaliefficientdocumentretrieval} as our retriever model. The training pipeline consists of two stages, both utilizing subsets of the Wiki-VISA dataset, while the other two datasets are reserved for out-of-domain (OOD) evaluation. 

We first conduct Supervised Fine-Tuning (SFT) on 2,417 curated samples to establish the basic citation capabilities. Then,
we  further optimize the agent policy via Group Relative Policy Optimization (GRPO) using a dataset of 1,000 samples to align the model with our citation-enhanced rewards objectives. Detailed hyperparameters and hardware configurations are provided in Appendix \ref{sec:appendix_implementation}.

\section{Experimental Results and Analysis}
\subsection{Main Experiment}

Table~\ref{tab:qwen-rag-results} compares MCite-RL with representative baselines on Wiki-VISA, FinRAGBench-V, and MMLongBench-Doc, evaluated in terms of Answer Accuracy (Ans) and Citation Precision (Cit). We summarize three key insights.

\paragraph{Multimodal citation remains challenging.}
As shown in Table~\ref{tab:qwen-rag-results}, vanilla RAG consistently yields low citation scores across datasets, indicating that coarse retrieval without explicit visual cropping is insufficient for citation tasks. Moreover, introducing agentic reasoning with iterative retrieval and cropping alone does not resolve this limitation. For instance, GPT-4o achieves only 3.40\% citation accuracy on Wiki-VISA, while Qwen2.5-VL-7B-Instruct attains merely 0.04\%. This implies that inference-time optimization without training is inadequate for reliable multimodal citation.

\paragraph{Citation-enhanced RL effectively improves citation precision.}

By explicitly incorporating citation-based rewards directly into the reinforcement learning objective, MCite-RL substantially improves citation performance across all datasets. Specifically, on Qwen2.5-VL-7B-Instruct backbone, it leads to double-digit improvements (over 10\%) compared to previous strong baselines, demonstrating the effectiveness of our citation-enhanced RL.

\paragraph{Improving citation precision generally leads to better answer accuracy.}

In most settings, citation improvements are accompanied by gains in answer accuracy. For Qwen2.5-VL-7B-Instruct, MCite-RL achieves the best answer precision in all datasets, including 60.00\% in Wiki-VISA, 21.81\% in FinRAGBench-V, and 35.04\% in MMLongBench-Doc, which corresponds to a 4–6\% improvement. For Qwen2.5-VL-3B-Instruct, however, clear gains in citation fail to yield corresponding answer accuracy improvements in some cases. This may imply that the limited capacity of the model restricts the generalization of citation benefits to the accuracy of the answer.

 Overall, these results indicate that explicitly optimizing citation  precision at the objective level is critical for reliable multimodal RAG, and that citation-enhanced reinforcement learning provides a systematic solution to this challenge. We further provide case study of our method in Appendix~\ref{sec:case_study}.
 
\begin{figure*}[t]
\centering
\setlength{\tabcolsep}{2pt}  

\begin{tabular}{cccc}
\begin{subfigure}[t]{0.23\textwidth}
    \centering
    \includegraphics[width=\linewidth]{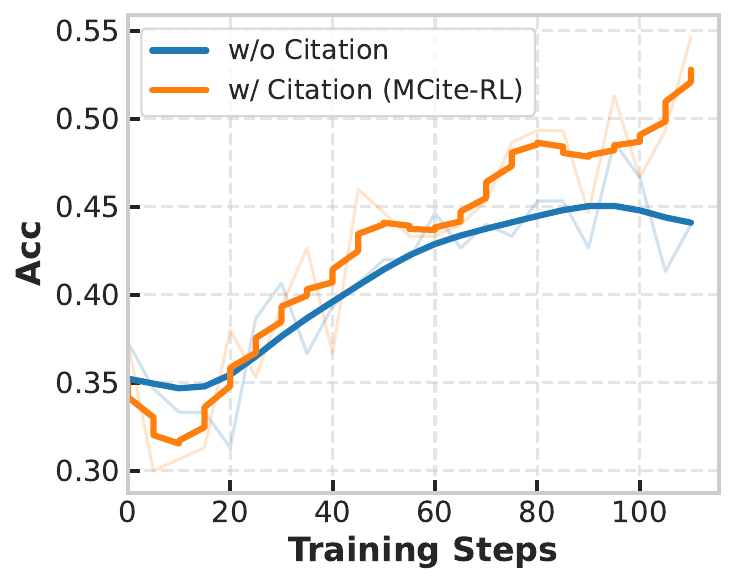}
    \caption{Answer Accuracy}
\end{subfigure}
&
\begin{subfigure}[t]{0.23\textwidth}
    \centering
    \includegraphics[width=\linewidth]{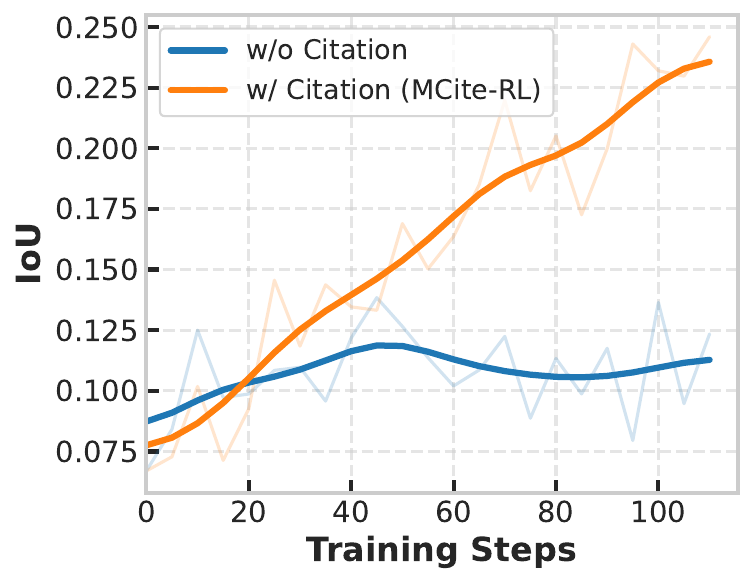}
    \caption{Outcome-level Citation}
\end{subfigure}
&
\begin{subfigure}[t]{0.23\textwidth}
    \centering
    \includegraphics[width=\linewidth]{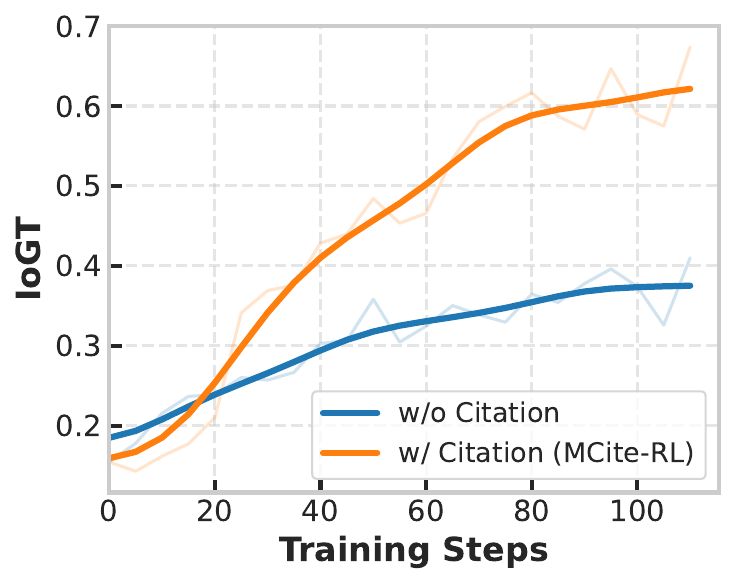}
    \caption{Process-level Citation}
\end{subfigure}
&
\begin{subfigure}[t]{0.23\textwidth}
    \centering
    \includegraphics[width=\linewidth]{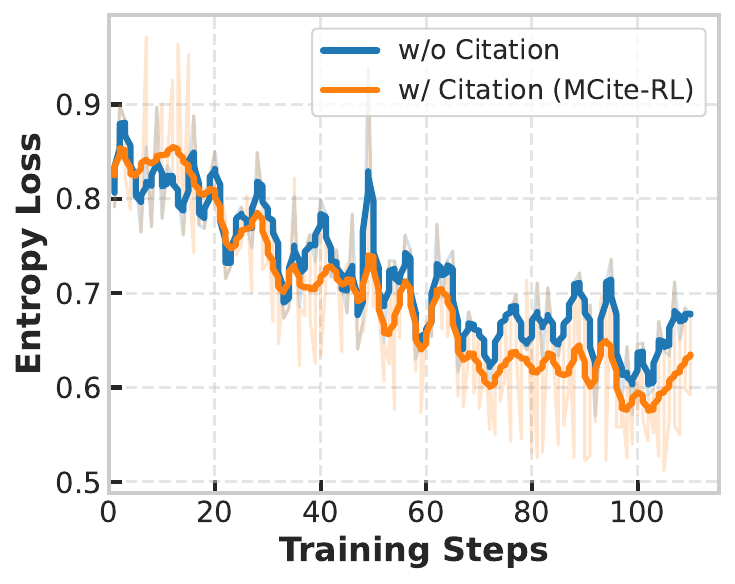}
    \caption{Entropy Loss}
\end{subfigure}

\end{tabular}

\caption{
\textbf{Training dynamics comparison across four key metrics.}
The orange solid lines represent MCite-RL, while the blue solid lines denote the baseline VRAG-RL (w/o Citation Reward).
}
\label{fig:four_metrics_comparison}
\end{figure*}
\subsection{Training Analysis}

To investigate the mechanisms driving the performance improvements, we analyze the training dynamics of MCite-RL against the highly competitive baseline (VRAG-RL w/o Citation), as illustrated in Figure~\ref{fig:four_metrics_comparison}. We derive two additional observations regarding the learning process. This analysis provides further insights into how citation-enhanced RL influences the learning behavior.

\paragraph{Citation rewards reduce generation uncertainty and improves overall quality.} The \textit{Entropy Loss} curve (Figure~\ref{fig:four_metrics_comparison} (d)) shows that although both models reduce entropy during training, MCite-RL consistently converges to a lower entropy level. This indicates that citation constraints serve as an effective regularizer, narrowing the feasible solution space and yielding more confident and robust generation.

\paragraph{Explicit citation rewards amplify the intrinsic correlation between answering and citation.}
In Figure~\ref{fig:four_metrics_comparison} (b) and (c), baseline curves (blue lines) reveal a slight natural increase in citation precision even without explicit citation supervision, suggesting an implicit link between answering and visual grounding. However, this signal remains weak and plateaus early, especially on out-of-domain datasets. In contrast, MCite-RL (orange lines), equipped with citation rewards, transforms this weak correlation into an explicit reinforcement loop, improving performance in both in-domain and out-of-domain scenarios. 


\subsection{Ablation and Citation Analysis}
\label{subsec:ablation}

To disentangle the contributions of different components and address potential entanglement between multi-term rewards, we conduct a component-wise ablation on the Qwen2.5-VL-7B-Instruct backbone (Table~\ref{tab:ablation_causal}). We further evaluate whether the generated citations provide semantically sufficient evidence for the answers through an answer-citation consistency analysis. 

\paragraph{Role of Training Stages.} SFT establishes the foundational agentic workflow and mitigates the cold-start challenge. However, removing the RL stage (\textit{w/o RL}) results in significant degradation in both answer and citation quality. This confirms that strategic reward optimization, rather than mere instruction following or format adherence, is the primary driver of MCite-RL’s performance gains.
\begin{table}[t]
\centering
\renewcommand{\arraystretch}{1.1}
\small
\begin{tabular}{l cc}
\toprule
\textbf{Configuration} & \textbf{Ans. (\%)} & \textbf{Cit. (\%)} \\
\midrule
\textbf{MCite-RL (Full)} & \textbf{60.00} & \textbf{36.05} \\ 
\midrule
\rowcolor{gray!5} \textit{Reward Components} & & \\
\quad w/o Process-level Cit.  & 57.60 & 34.50 \\
\quad w/o Outcome-level Cit.  & 54.00 & 13.15  \\
\quad w/o All Cit. Rewards    & 54.20 & 20.56 \\
\quad w/o Ans. Reward         & 53.20 & 41.65  \\
\quad w/o Search Reward       & 54.80 & 35.95 \\
\quad w/o Format Constraints  & 55.20 & 33.41 \\
\midrule
\rowcolor{gray!5} \textit{Training Ablation} & & \\
\quad w/o RL Stage (SFT only) & 45.80 & 18.66 \\
\quad w/o Training (Base)     & 9.60  & 0.04  \\
\bottomrule
\end{tabular}
\caption{Component-wise Ablation for Reward Function on Wiki-VISA (7B). We categorize components into reward items and training constraints to analyze their contributions.}
\label{tab:ablation_causal}
\end{table}
\paragraph{Effectiveness of Visual Citation Rewards.} Our ablation results (Table~\ref{tab:ablation_causal}) demonstrate the significant impact of visual citation-enhanced rewards on overall system performance. Removing all visual citation rewards results in a sharp decline in citation precision from 36.05\% to 20.56\%, while answer accuracy concurrently drops from 60.00\% to 54.20\%. This confirms that visual citation rewards not only enhance localization quality but also improve answer generation through evidence-based reasoning. Within this framework, the outcome-level reward serves as the dominant anchor by providing precise IoU feedback, while the process-level reward provides additional refinement by constraining intermediate cropping. Crucially, without outcome-level anchors (\textit{w/o Outcome-level Vis-Cit.}), the model tends to generate overly expansive crops to satisfy intermediate coverage signals at the expense of precision, leading to a substantial degradation in citation quality (13.15\%). Ultimately, the synergy between both reward levels is essential for maintaining precise, stable, and verifiable agentic visual citation in multimodal RAG.

\paragraph{Answer–Citation Consistency Evaluation.} Since bounding-box IoU only measures spatial alignment, we further evaluate whether the cited regions semantically support the generated answers. We adopt an LLM-as-a-judge protocol that assesses whether the cited evidence is sufficient, non-redundant, and directly supports the answer. MCite-RL consistently improves this metric over baselines, indicating better semantic alignment between generated answers and visual citations (detailed results in Appendix~\ref{sec:answer_citation_consistency} Table~\ref{tab:answer_citation_consistency}).

\section{Conclusion}
Accurate citation is essential for reliable multimodal RAG systems, yet it remains challenging in practice. In this work, we propose MCite-RL, a framework that integrates agentic workflow with a citation-enhanced reward function. By explicitly supervising citation at both the process and outcome levels, MCite-RL enables joint optimization of citation and answer quality.

\section*{Limitations}
While MCite-RL improves the precision of visual citation through citation-aware reinforcement learning, it has several limitations. The current training process introduces some additional computational overhead, and the formulation focuses on bounding-box–based visual citation on single images and does not explicitly model more expressive citation structures; we leave these extensions for future work.
\section*{Ethical Consideration}
This work focuses on improving the traceability and verifiability of multimodal retrieval-augmented generation through explicit visual citation. By encouraging models to ground their outputs in identifiable visual evidence, our approach aims to reduce hallucination and improve transparency. These goals are broadly aligned with responsible AI practices. Nevertheless, the system may inherit biases present in the underlying datasets, including those related to visual representations and annotation practices, which could potentially affecting citation behavior. In addition, while visual citation enhances interpretability, it does not guarantee factual correctness of the retrieved content itself. Caution should be exercised when deploying such systems in high-stakes domains, and human oversight remains necessary.

\section*{Acknowledgement}
We thank the anonymous reviewers for their helpful
comments on this paper. This work was partially
supported by National Natural Science Foundation of China projects (No. 62476010), National Natural
Science Foundation of China (No. 62272008), and the Fundamental Research Funds for the Central Universities, Peking University.
\bibliography{custom}
\clearpage
\appendix

\section{Additional Experimental Analysis}
\subsection{Answer-Citation Consistency Evaluation}\label{sec:answer_citation_consistency}
IoU-based citation metrics primarily measure spatial alignment between predicted and ground-truth regions, but do not directly assess whether the cited region semantically supports the generated answer. To address this limitation, we introduce an additional answer-citation consistency evaluation based on an LLM-as-a-judge paradigm. Given the cropped citation region, question, and generated answer, the evaluator judges whether the citation provides sufficient, non-redundant, and directly supporting evidence.
\begin{table}[h]
\centering
\resizebox{\columnwidth}{!}{
\begin{tabular}{lccc}
\toprule
\textbf{Method} & \textbf{Wiki-VISA} & \textbf{FinRAGBench-V} & \textbf{MMLongBench-Doc} \\
\midrule
GPT-4o & 17.00 & 27.59 & 26.56 \\
Gemini-2.0-flash & 0.60 & 12.20 & 6.22 \\
Claude-4.5-sonnet & 2.20 & 28.15 & 26.79 \\
\midrule
\multicolumn{4}{c}{\textbf{Qwen2.5-VL-3B-Instruct}} \\
\midrule
Vanilla RAG & 1.20 & 0.29 & 2.63 \\
ReAct & 0.00 & 0.00 & 0.00 \\
Search-R1 & 5.80 & 0.07 & 0.96 \\
VRAG-RL (w/SFT) & 23.40 & 18.39 & 26.32 \\
MCite-RL & \textbf{44.00} & \textbf{32.95} & \textbf{38.76} \\
-- w/o Proc & 37.80 & 20.33 & 29.19 \\
-- w/o RL & 20.40 & 10.20 & 16.99 \\
-- w/o Training & 0.00 & 2.89 & 1.20 \\
\midrule
\multicolumn{4}{c}{\textbf{Qwen2.5-VL-7B-Instruct}} \\
\midrule
Vanilla RAG & 3.40 & 6.47 & 16.03 \\
ReAct & 3.20 & 0.79 & 5.02 \\
Search-R1 & 5.00 & 1.51 & 1.44 \\
VRAG-RL (w/SFT) & 22.80 & 12.43 & 18.73 \\
MCite-RL & \textbf{44.80} & \textbf{44.83} & \textbf{49.26} \\
-- w/o Proc & 37.20 & 25.14 & 34.63 \\
-- w/o RL & 22.00 & 20.89 & 22.01 \\
-- w/o Training & 0.80 & 4.36 & 0.96 \\
\bottomrule
\end{tabular}
}
\caption{Answer-citation consistency (\%) of MCite-RL compared with other methods. Higher scores indicate better semantic alignment between generated answers and cited regions.}
\label{tab:answer_citation_consistency}
\end{table}

As shown in Table~\ref{tab:answer_citation_consistency}, MCite-RL consistently outperforms baselines across all three datasets under this evaluation, demonstrating that it not only localizes relevant regions more precisely but also identifies evidence that is more directly supportive of the generated answers.
\subsection{Case Study}\label{sec:case_study}
\begin{figure*}[t]
  \centering
  \includegraphics[clip=true,width=0.9\textwidth]{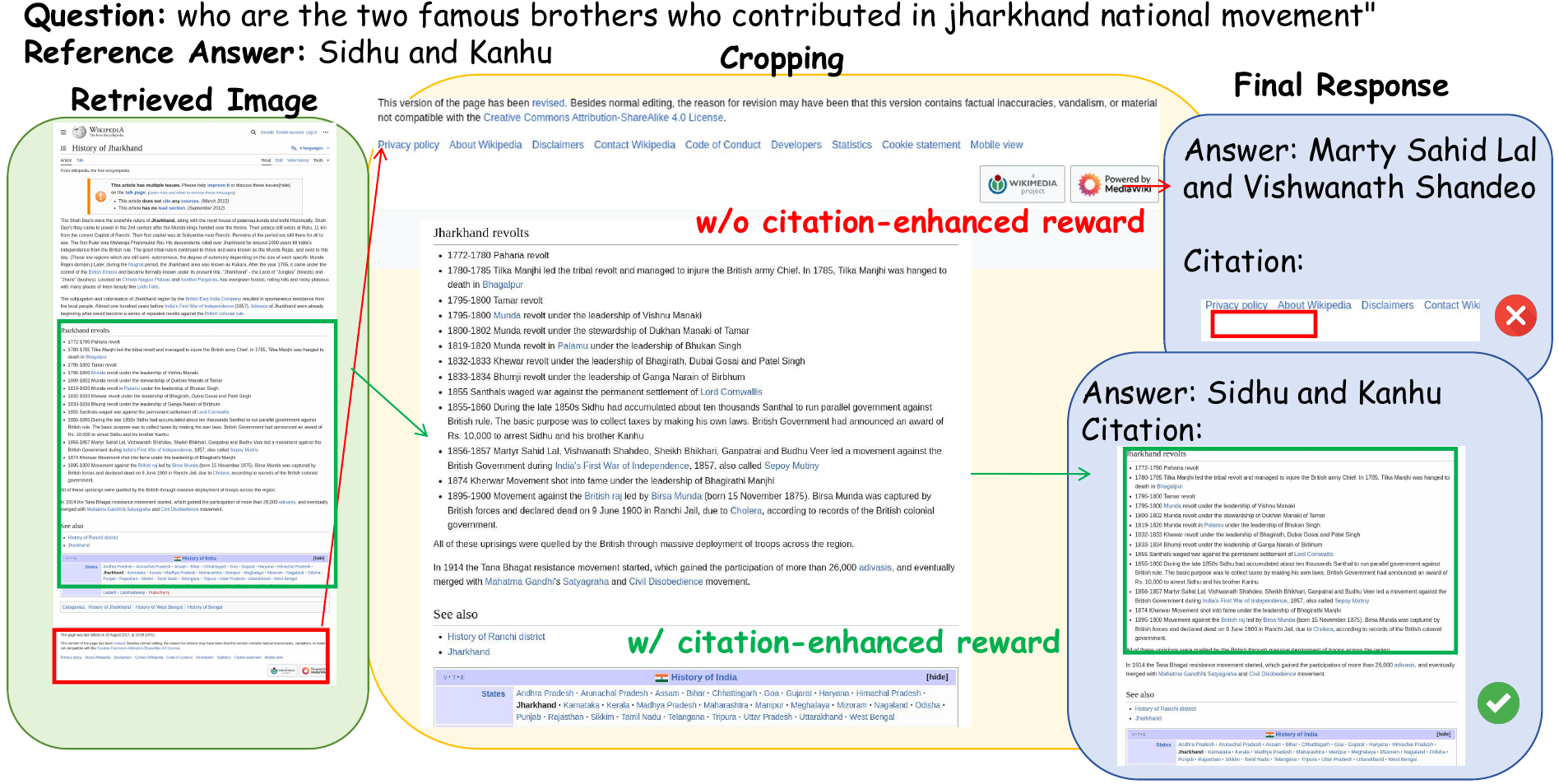}
  \caption{This is a qualitative comparison showing the impact of the Citation-enhanced Reward. Without the reward (top), incorrect cropping leads to information loss and hallucinations; with the reward (bottom), the model precisely locates critical evidence to generate accurate answers and supporting citations.}
  \label{fig:case_study}
\end{figure*}

To vividly illustrate the effectiveness of our proposed method, Figure~\ref{fig:case_study} presents a qualitative comparison between models with and without citation-based rewards on a query regarding the ``Jharkhand national movement''. Without the citation-enhanced reward, the model fails to localize the relevant evidence, instead cropping an irrelevant footer region, which consequently leads to the incorrect answer ``Marty Sahid Lal and Vishwanath Shandeo''. In contrast, with the citation-enhanced reward, our MCite-RL method effectively guides the model to precisely localize the key ``Jharkhand revolts'' paragraph. This accurate evidence localization enables the model to correctly identify the answer ``Sidhu and Kanhu'', supported by an accurate visual citation, thereby demonstrating the effectiveness of our proposed method.

\section{Details for SFT Trajectory Synthesis and Filtering}\label{appendix:SFT_DATA}

To construct effective process supervision for the agentic SFT stage, we synthesize trajectories following a structured search--crop--answer workflow. Each trajectory contains intermediate retrieval and visual grounding actions, followed by the final answer with visual citation. To ensure that the synthesized trajectories provide reliable and learnable supervision signals, we further apply strict yet practical filtering criteria to remove noisy or inconsistent samples.

\subsection{Trajectory Synthesis}

We synthesize agentic trajectories using a multimodal teacher model. Given a question and the corresponding document image, the teacher is instructed to first retrieve relevant evidence, progressively refine the visual region through hierarchical cropping, and finally produce the answer with the corresponding citation bounding box.

The generation process follows the structured workflow shown in Table~\ref{tab:trajectory_prompt}, which specifies the available tools, search-crop refinement strategy, and required output format. This design encourages trajectories with explicit intermediate evidence localization steps, providing process-level supervision for agentic SFT.

\begin{table*}[t]
\centering
\resizebox{\textwidth}{!}{
\begin{tabular}{p{0.18\textwidth} p{0.75\textwidth}}
\toprule
\textbf{Component} & \textbf{Instruction} \\
\midrule

Character Introduction &
You are an intelligent assistant that performs search, visual grounding, and precise question answering. 
You must reason step by step and use tools when necessary. 
If the answer is uncertain, use the search tool. If a specific image region is needed, use the crop tool.
\\

\midrule

Available Tools &
\textbf{search}: Retrieve relevant information using keywords or questions. \newline
\textbf{crop}: Crop a region from the image. \newline
\textbf{answer}: Provide the final answer with a visual citation (bounding box).
\\

\midrule

Workflow &
\textbf{Step 1: SEARCH} \newline
- You MUST begin with a search. \newline
- Do NOT skip this step. \newline

\textbf{Step 2: CROP} \newline
- Progressively crop the region after search. \newline
- The first crop should be a large semantic region and must fully contain the ground-truth citation region. \newline
- Subsequent crops should become smaller and more focused. \newline
- The final crop should still contain the citation region and be close to, but not identical to, the final citation.
\\

\midrule

Iterative Refinement &
After each step, evaluate whether the information is sufficient. \newline
If sufficient, continue refining the current region until it is precise enough to support the answer. \newline
If insufficient, perform a new search with refined keywords and restart refinement from a larger region.
\\

\midrule

Refinement Strategy &
Follow a hierarchical zoom-in process:
Large region $\rightarrow$ Intermediate region $\rightarrow$ Near-final region.
\\

\midrule

Constraints &
- Do NOT over-crop irrelevant images. \newline
- Early crops MUST NOT match or approximate the final citation. \newline
- Each crop MUST be smaller and more focused than the previous one. \newline
- The final citation MUST EXACTLY match ground-truth coordinates.
\\

\midrule

Final Answer &
Only answer after completing search and crop steps. \newline
Provide the final answer and exact citation bounding box.
\\

\midrule

Reply Format &
\textbf{SEARCH}: $\{\texttt{think},\texttt{search}\}$ \newline
\textbf{CROP}: $\{\texttt{think},\texttt{bbox},\texttt{description}\}$ \newline
\textbf{Final Answer}: $\{\texttt{think},\texttt{answer},\texttt{citation}\}$
\\

\bottomrule
\end{tabular}
}
\caption{Trajectory generation prompt used for synthesizing agentic SFT trajectories.}
\label{tab:trajectory_prompt}
\end{table*}

\subsection{Trajectory Filtering}

Although the teacher model can generate diverse trajectories, some samples may contain incorrect retrieval decisions, inaccurate answers, or invalid citation regions. Therefore, we apply a set of filtering criteria (Table~\ref{tab:filtering_criteria}) to ensure the quality of the synthesized supervision data.

Specifically, we filter trajectories based on retrieval correctness, answer correctness, and bounding box validity. Only trajectories satisfying all quality constraints are retained for SFT training. In total, we generate 28,175 candidate trajectories and retain 2,417 trajectories after filtering, resulting in an acceptance rate of approximately 8.6\%. Table~\ref{tab:trajectory_sample} shows an example of filtered trajectory.
\begin{table*}[h]
    \centering
    \small
    \begin{tabularx}{\linewidth}{X}
        \toprule
        \textbf{Criteria for Trajectory Quality Control} \\
        \midrule
        A trajectory is retained only if the MLLM:\\
        (i) Successfully retrieves the correct image, ensuring relevant visual context is available; \\
        \addlinespace
        (ii) Generates the correct answer, guaranteeing semantic correctness; \\
        \addlinespace
        (iii) Predicts a valid citation box with high overlap with the ground-truth citation region;\\
        \addlinespace
        (iv) Ensures the first crop fully contains the ground-truth citation box, providing a valid starting region for subsequent iterative focusing. \\
        \bottomrule
    \end{tabularx}
     \caption{Filtering criteria for trajectory synthesis. A trajectory is retained only if it satisfies all the conditions.}
    \label{tab:filtering_criteria}
\end{table*}

\section{Visual Citation Annotation Guidelines}
\label{subsec:annotation_guidelines}

To evaluate the quality of visual citation on MMLongBench-Doc, we developed a strict annotation protocol to annotate 300 samples. Annotators are provided with a triplet: \texttt{(User Query, Correct Answer, Source Image)}. The task is to draw a bounding box that visually substantiates the answer. The guidance is listed in Table \ref{tab:simple_guidelines}. We have two annotators to double check each data.

\begin{table*}[t] 
    \centering
    \small 
    \renewcommand{\arraystretch}{1.3} 
    
    \begin{tabularx}{\textwidth}{p{0.1\textwidth} X}
        \toprule
        \textbf{Role} & \textbf{Trajectory Content (Reasoning, Action, and Observation)} \\
        \midrule
        
        \textbf{User} & \textbf{Question:} who played the preacher in the movie uncle drew \\
        \midrule
        
        \textbf{Model} & \textcolor{gray}{\texttt{<think>}} I need to find an image that identifies which actor played the character Preacher in the movie Uncle Drew. I'll start by searching for an image of the cast or a relevant listing that mentions 'Preacher' and the actor's name. \textcolor{gray}{\texttt{</think>}} \newline
        \textbf{Action:} \texttt{<search>Uncle Drew Preacher actor</search>} \\
        \midrule
        
        \textbf{Environment} & \textit{[System returns retrieved full-page screenshot]} \\
        \midrule
        
        \textbf{Model} & \textcolor{gray}{\texttt{<think>}} The Wikipedia page likely contains the cast list where the character Preacher is mentioned. I will start with a large crop over the lower section to see if any cast or relevant lines mention 'Preacher', then progressively narrow down. \textcolor{gray}{\texttt{</think>}} \newline
        \textbf{Action:} \texttt{<bbox>[22, 895, 531, 1119]</bbox>} \textit{(Coarse Crop)} \\
        \midrule
        
        \textbf{Environment} & \textit{[System returns coarse cropped image]} \\
        \midrule
        
        \textbf{Model} & \textcolor{gray}{\texttt{<think>}} The first crop shows the 'Cast' section... I need to zoom in further to the specific line mentioning 'Preacher' to identify the actor who played that role. \textcolor{gray}{\texttt{</think>}} \newline
        \textbf{Action:} \texttt{<bbox>[6, 257, 1164, 412]</bbox>} \textit{(Fine Crop)} \\
        \midrule
        
        \textbf{Environment} & \textit{[System returns fine cropped image]} \\
        \midrule
        
        \textbf{Model} & \textcolor{gray}{\texttt{<think>}} The cropped Wikipedia cast section includes the line specifying who played Preacher. It states 'Chris Webber as Preacher.' I will provide the final answer... \textcolor{gray}{\texttt{</think>}} \newline
        \textbf{Final Response:} \texttt{<answer>Chris Webber</answer>} \newline
        \textbf{Citation:} \texttt{<citation>[9, 123, 2117, 185]</citation>} \\
        
        \bottomrule
        \multicolumn{2}{p{\dimexpr\textwidth-2\tabcolsep}}{
            \vspace{0.5em}
            \textbf{Sample Meta-data:} \texttt{EM\_Score: 1.0} | \texttt{Search: Success} | \texttt{Citation: Consistent} | \texttt{BBox\_Grounding: Correct} \newline
            \textit{Note: The citation bbox coordinates are relative to the final cropped image input.}
        }
    \end{tabularx}
    \caption{\textbf{Synthesized Trajectory Data Sample.} This table visualizes a complete multi-turn interaction where the model performs search and progressive cropping to answer the user query. Note how the bounding box coordinates refine from coarse to fine.}
    \label{tab:trajectory_sample}
\end{table*}

\begin{table*}[h]
    \centering
    \small 
    \renewcommand{\arraystretch}{1.5}
    \begin{tabular}{p{0.12\textwidth} p{0.50\textwidth} p{0.30\textwidth}}
        \toprule
        \textbf{Principle} & \textbf{Guideline} & \textbf{Avoid (What to reject)} \\
        \midrule
        
        \textbf{1. Completeness} & 
        The bounding box must fully encompass the visual evidence (text span or data row) required to answer the query. & 
        Cutting off key context or truncated sentences. \\
        
        \textbf{2. Compactness} & 
        The box should be the \textit{smallest} possible rectangle that encloses the evidence. It must exclude irrelevant surrounding sentences or whitespace. & 
        Including the whole paragraph or page unnecessarily. \\
        
        \textbf{3. Table Logic} & 
        For structured data (tables/lists), annotate the specific row(s) containing both the entity and the value to preserve semantic meaning. & 
        Selecting only the value cell without row headers. \\
        
        \textbf{4. Relevance} & 
        If the answer appears in multiple locations, select the single most informative region (e.g., the main definition rather than a summary). & 
        Annotating multiple duplicate regions or ambiguous footnotes. \\
        
        \bottomrule
    \end{tabular}
    \caption{\textbf{Annotation Principles for Visual Citation.} The ground truth bounding box must balance completeness, compactness, and relevance.}
    \label{tab:simple_guidelines}
\end{table*}

\section{Implementation Details}
\label{sec:appendix_implementation}

\subsection{Hardware and Software Environment}
\label{subsec:hardware}
All experiments were conducted on a high-performance computing cluster, using GPU: 8*NVIDIA A800 (80GB), and vLLM as the inference engine. SFT takes about 4 hours, and RL completes in approximately 8 hours under our current configuration (1,000 samples, 1 epoch).

\subsection{Reward Configuration}
The composite reward function aims to balance retrieval quality, reasoning correctness, and visual alignment. The specific weights for each component are listed in Table \ref{tab:reward_config}. We enforce a format pattern constraint: if the model output fails to match the required structural pattern, the total reward is penalized to 0.
\begin{table}[h]
    \centering
    \begin{tabular}{l c c}
        \toprule
        \textbf{Reward Component} & \textbf{Metric} & \textbf{Weight}\\
        \midrule
        Retrieval Reward & NDCG & 0.15 \\
        Answer Reward & Exact Match & 0.40 \\
        Citation Outcome Reward & IoU & 0.35 \\
        Citation Process Reward & IoGT & 0.10 \\
        \bottomrule
    \end{tabular}
    \caption{Reward components, corresponding metrics, and weights used in RL optimization.}
    \label{tab:reward_config}
\end{table}

\subsection{Training Hyperparameters}
We conducted a two-stage training process: Supervised Fine-Tuning (SFT) followed by Reinforcement Learning (RL).

\textbf{Supervised Fine-Tuning (SFT).} 
We fine-tuned the Qwen2.5-VL-7B-Instruct model using the configuration detailed in Table \ref{tab:sft_params}.

\begin{table}[h]
    \centering
    \small
    \begin{tabular}{l c}
        \toprule
        \textbf{Hyperparameter} & \textbf{Value} \\
        \midrule
        Base Model & Qwen2.5-VL-7B-Instruct \\
        Training Epochs & 3 \\
        Learning Rate & $1.0 \times 10^{-5}$ \\
        LR Scheduler & Cosine (Warmup ratio 0.1) \\
        Global Batch Size & 16 \\
        Gradient Accumulation & 2 \\
        Max Context Length & 16,384 \\
        Training Samples & 2417\\
        \bottomrule
    \end{tabular}
     \caption{Hyperparameters for SFT}
    \label{tab:sft_params}
\end{table}

\textbf{Reinforcement Learning (RL).}
In the RL stage, we utilized the GRPO algorithm. The detailed hyperparameters are provided in Table \ref{tab:rl_params}.

\begin{table}[h]
    \centering
    \small
    \begin{tabular}{l c}
        \toprule
        \textbf{Hyperparameter} & \textbf{Value} \\
        \midrule
        Algorithm & GRPO \\
        Training Epochs & 1 \\
        Actor Learning Rate & $1.0 \times 10^{-6}$ \\
        KL Coefficient & 0.01 \\
        Global Train Batch Size & 8 \\
        Max Prompt Length & 8,192 \\
        Max Response Length & 2,048 \\
        Rollout ($N$) & 1 \\
        Number of Agents & 5 \\
        Training Samples & 1000\\
        \bottomrule
    \end{tabular}
    \caption{Hyperparameters for RL}
    \label{tab:rl_params}
\end{table}
\subsection{Dataset Statistics}
\label{subsec:data_stats}
We use the subset of Wiki-Visa dataset for training, and subsets of FinRAGBench-V and MMLongBench-Doc as the OOD evaluation sets. The retrieval corpus consists of the source documents from which the model retrieves information. The detailed statistics are shown in Table \ref{tab:data_stats}.

\begin{table}[h]
    \centering
    \small
    \begin{tabular}{l c c}
        \toprule
        \textbf{Dataset} & \textbf{Retrieval Corpus} & \textbf{QA Pairs} \\
        \midrule
        Wiki-VISA & 12,387& 500 \\
        FinRAGBench-V & 3,011 & 1,398 \\
        MMLongBench-Doc & 6,527 & 418 \\
        \bottomrule
    \end{tabular}
    \caption{Data Statistics of the specific retrieval corpus size and validation set size for each dataset.}
    \label{tab:data_stats}
\end{table}

\subsection{Details of Evaluation Metrics}
\label{appendix:metrics}
To ensure the reproducibility of our experiments, we follow the official evaluation protocols for each benchmark. For Wiki-VISA and MMLongBench-Doc, we adopt Relaxed Exact Match (EM) and ROUGE-L, as their answers are predominantly short and fact-based with limited lexical variation. In contrast, for FinRAGBench-V, which contains a mixture of short factual and long-form reasoning responses, we employ a GPT-4o-based judge for semantic evaluation. These settings align with the specific characteristics of each dataset.

\end{document}